\documentclass[letterpaper, 10 pt, conference]{ieeeconf}  

\IEEEoverridecommandlockouts                              

\usepackage{graphicx}
\usepackage{amsmath} 
\usepackage[dvipsnames]{xcolor}
\usepackage{hyperref}

\title{\LARGE \bf \textbf{T-FunS3D}:
\textbf{T}ask-Driven Hierarchical Open-Vocabulary \\\textbf{3D} \textbf{Fun}ctionality \textbf{S}egmentation
}

\author{Jingkun Feng$^{1}$ and Reza Sabzevari$^{1}$
\thanks{$^{1}$Jingkun Feng and Reza Sabzevari are with the P4MARS Lab at the Faculty of Aerospace Engineering, Delft University of Technology.}%
}

\begin{document}
\maketitle
\thispagestyle{empty}
\pagestyle{empty}

\begin{abstract}

Open-vocabulary 3D functionality segmentation enables robots to localize functional object components in 3D scenes. It is a challenging task that requires spatial understanding and task interpretation. 
Current open-vocabulary 3D segmentation methods primarily focus on object-level recognition, while scene-wide part segmentation methods attempt to segment the entire scene exhaustively, making them highly resource-intensive and time consuming.
Balancing segmentation performance in terms of granularity, accuracy, and speed remains a challenge.
As one step towards alleviating this, we introduce T-FunS3D, a task-driven hierarchical open-vocabulary 3D functionality segmentation method that provides actionable perception for robotic applications.
Our method takes as input the 3D point cloud and posed RGB-D images of an indoor scene.
We construct an open-vocabulary scene graph by extracting instances and their visual embeddings in the environment.
Given a task description, T-FunS3D identifies the most relevant instances in the scene graph and locates their functional components leveraging a vision-language model.
Experiments on the SceneFun3D dataset demonstrate that \text{T-FunS3D} is comparable to state-of-the-art in open-vocabulary 3D functionality segmentation, while achieving faster runtime and reduced memory usage.

\textnormal{\textit{Supplementary materials and code: \href{https://t-funs3d.github.io}{t-funs3d.github.io}.}}
\end{abstract}

\section{INTRODUCTION}

Open-vocabulary 3D functionality segmentation extracts the semantic meaning and location of functional interactive elements in 3D scenes without relying on a fixed set of predefined annotations.
This capability provides robots with both concrete objects and their functional parts, enabling them to execute the assigned tasks.
Task-driven functional object segmentation, also referred to as task-driven affordance grounding in \cite{delitzas2024scenefun3d}, predicts the masks associated to Gibsonian affordances~\cite{gibson2014ecological} inferred from telic affordances~\cite{pustejovsky1991generative} given as free-form task descriptions.
It poses a unique challenge that requires both the physical understanding of the functionalities of objects and the interpretation of complex linguistic expressions.
To tackle this challenge, this paper proposes a novel pipeline for hierarchical scene understanding and functionality segmentation driven by tasks using an open-vocabulary scene graph and vision-language models~(VLMs).

\begin{figure}[ht]
\centering
\includegraphics[width=\linewidth]{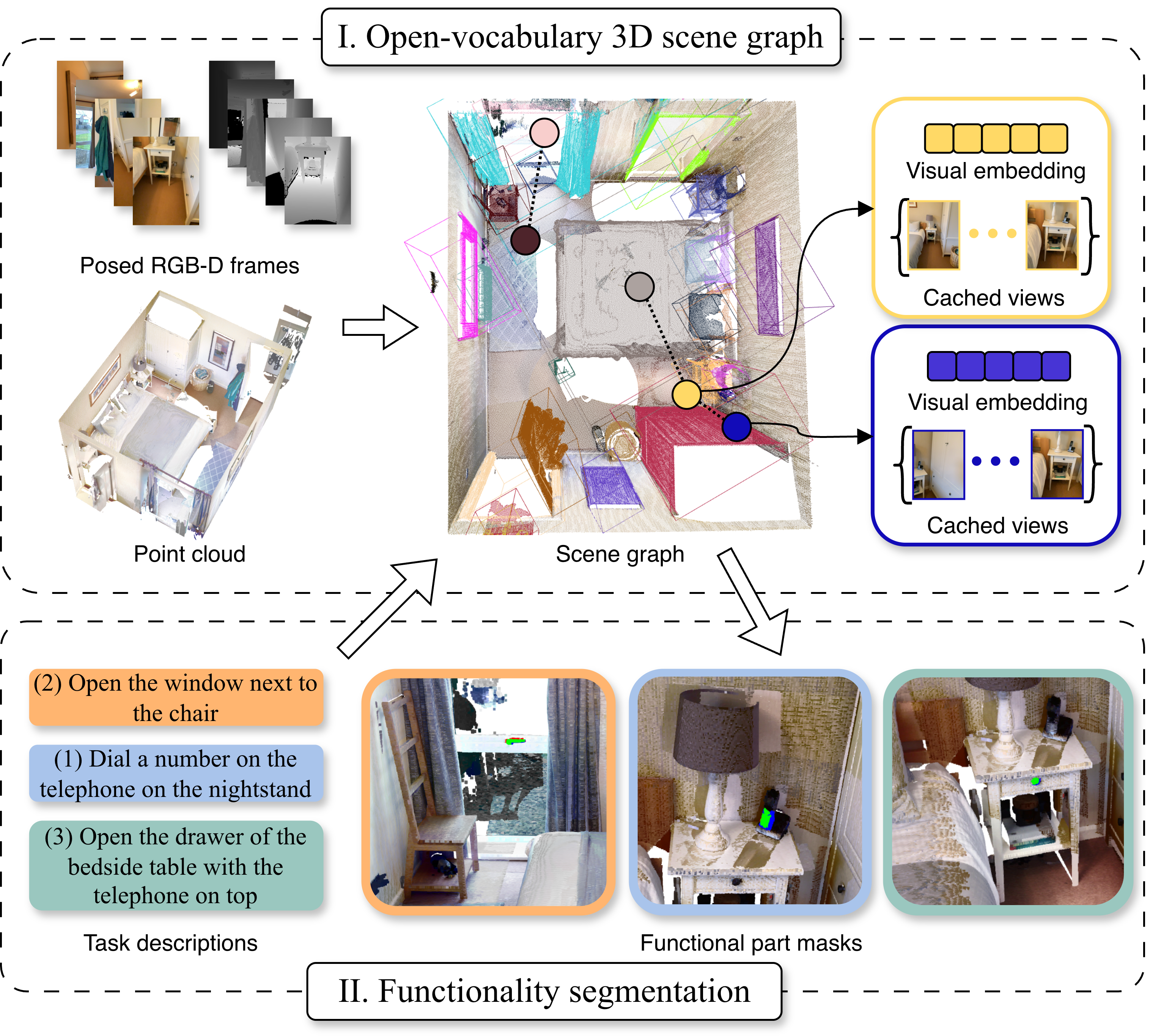}
\caption{T-FunS3D: a training-free method for open-vocabulary functionality segmentation in 3D scenes based on free-form text descriptions. 
Taking as input a point cloud of an indoor environment and the corresponding posed RGB-D images, an open-vocabulary scene graph is constructed at stage I, which contains instance embeddings and inter-object relation embeddings. 
At stage II, once a task is assigned, T-FunS3D extracts the functional components required to undertake the interaction from the scene.} 
\vspace{-.6cm}
\label{fig:teaser}
\centering
\end{figure}
Conventional 3D semantic understanding methods are primarily trained on labeled datasets to recognize a considerable number of object classes.
Using large-scale annotations, these models work well for close-set instance segmentation.
However, because they are trained in a supervised or semi-supervised manner, they fall short of generalizing to novel object classes.
In contrast, open-vocabulary methods leverage large language models to infer scene semantics, supporting 3D exploration guided by free-form text descriptions.

While most existing methods focus on recognizing object-level instances, only a few particularly address finer-grained segmentation within scenes~\cite{takmaz_search3d_2024, corsetti_functionality_2024, zhang2025open, rotondi_fungraph_2025}, including functional object parts, such as a door handle.
However, such fine-grained entities are crucial for downstream tasks like scene interaction.
For example, assistive robots must recognize not only objects in an open-set world but also their fine-grained components, such as the chair arm and door of a washing machine, to execute general interactive tasks.
Previous work on this problem often performs fine-grained segmentation throughout the scene without task-specific distinctions~\cite{takmaz_search3d_2024, yang2023sam3d}. 
Nonetheless, the methodological design they followed is resource-intensive, imposing high computational overhead and memory costs, which are not accessible in most mobile robotic systems.
Therefore, there is a pressing need for more efficient designs that extract fine-grained open-vocabulary segmentation in a task-specific manner.

Hierarchical open-vocabulary scene understanding methods have recently attracted growing interest~\cite{werby2024hierarchical, maggio2024clio, takmaz_search3d_2024} and demonstrated promising results.
The majority of them operates at higher levels of abstraction from region to object.
They provide no identification of the fine-grained functional details of scene entities for interactive robot applications.
Additionally, methods capable of part-level segmentation such as Search3D~\cite{takmaz_search3d_2024} often over-segment the entire scene to detect every entity at various level of granularity.
However, in practice, only a small subset of objects in the scene is relevant for an assigned task, making scene-wide over-segmentation an unnecessary burden on computation and storage.

In light of these limitations, we advocate for an efficient segmentation approach focusing on task-relevant areas and performing functional part segmentation exclusively for selected objects.
Our main contributions are as follows:
\begin{itemize}
    \item We introduce a training-free method for 3D functionality segmentation based on free-form task descriptions, which leverages an actionable 3D scene graph containing open-vocabulary semantics.
    \item We present a task-driven hierarchical approach that first decomposes a scene into object instances, then segments fine-grained functional components from task-related entities, thereby facilitating efficient deployment in real-world robotic applications. 
    \item We propose to construct an open-vocabulary scene graph that encodes nodes and inter-object edges with visual embedding features, enabling effective localization of entities with reference to the surrounding environment in an open-world setting.
    \item Our method reduces the runtime and memory consumption for open-vocabulary 3D segmentation of functional object components on the SceneFun3D dataset~\cite{delitzas2024scenefun3d} while maintaining highly competitive accuracy.
\end{itemize}
\section{Related Work}

\subsection{Open-Vocabulary 3D Scene Understanding}
Recent works on open-vocabulary 3D scene understanding have made significant progress.
They typically focus on object-level semantic segmentation following two main streams.
One leverage 2D foundation models to decompose the observation and segment objects~\cite{takmaz_openmask3d_2023, nguyen_open3dis_2023, huang_openins3d_2023}. 
The other learns to embed semantics in explicit primitives such as points~\cite{jatavallabhula2023conceptfusion} and Gaussian ellipsoids~\cite{li2025scenesplat} or in implicit fields~\cite{kerr2023lerf, engelmann2024opennerf}.
However, they separate objects from scenes but do not understand relationships between entities.
Therefore, despite achieving promising instance segmentation, they remain bounded in the scope of open-world querying capabilities, struggle to handle complex reasoning.

To overcome the limitations of isolated instance representations, efforts have been given to understand and represent relationships between entities.
For instance, ConceptGrpahs~\cite{gu2024conceptgraphs}, OVSG~\cite{changcontext}, Open3DSG~\cite{koch_open3dsg_2024} constructs scene graphs with inter-object relationships.
Following another line, Locate3D~\cite{mcvaylocate} learns a contextualized point cloud to empower referential grounding.
However, they are largely restricted to the instance-level understanding at the finest level, and thus struggle to identify parts.
Additionally, they either rely on extensive use of large models or construct exhaustive contextualize point-based representations, imposing high computational and memory costs.

These limitations in prior work highlight a clear absence of a comprehensive method that seamlessly integrates part-level open-vocabulary scene understanding with efficient and actionable representation. 
To bridge this gap, our method constructs a lightweight open-vocabulary 3D scene graph that encodes both nodes and inter-object edges using visual embeddings.
We propose to interpret the scene in a hierarchical task-driven manner, which enables efficient extraction of fine-grained functional parts without expensive computational and memory overhead.

\subsection{3D Part Segmentation and Functionality Segmentation}
3D part segmentation has made some progress in open-vocabulary-based frameworks thanks to recent efforts in~\cite{takmaz_search3d_2024, liu2023partslip, zhou2023partslip++}.
Similar to open-set instance segmentation, a common strategy here is to lift features or predition from 2D foundation models (e.g., \cite{kirillov2023segment}, \cite{radford2021learning}, \cite{balibar2006glip}, \cite{caron2021DINO}) into 3D.
For instance, PartSLIP~\cite{liu2023partslip} and its successor \cite{zhou2023partslip++} reprojects language-guided open-vocabulary part detection on rendered images into 3D.
Similarly, SAMPart3D~\cite{yang_sampart3d_2024} distills 2D features into 3D representation.
But these methods remain limited to single-object 3D models, restricting their applicability to full-scene understanding.

As an extension of part segmentation, functionality segmentation, recently introduced in SceneFun3D~\cite{delitzas2024scenefun3d}, focuses on segmenting functional interactive sub-parts in 3D scenes according to free-form language descriptions of tasks.
Given its crucial role in downstream robotic applications like planning and manipulation, a growing body of research ~\cite{takmaz_search3d_2024,corsetti_functionality_2024, zhang2025open, rotondi_fungraph_2025} aims to address this challenge.

Conceptually, our work is closely related to Search3D~\cite{takmaz_search3d_2024} and Fun3DU~\cite{corsetti_functionality_2024}. 
Search3D segments point clouds and exhaustively embeds all objects and their parts.
Fun3DU, in turn, detects objects of interest across all images and lifts 2D segmentation masks of their functional parts into 3D.
However, it detects merely objects given in the query.
Consequently, it must re-execute the object detection process upon receiving any new query, making it highly inefficient.
Unlike these methods, our approach constructs an open-vocabulary scene graph for the entire 3D environment and retrieves functional parts exclusively from the selected objects based on the given task.
This eliminates repeated instance detection and enables efficient handling of complex 3D grounding tasks.

\begin{figure*}[htbp]
\centering
\vspace{.1cm}
\includegraphics[width=\linewidth]{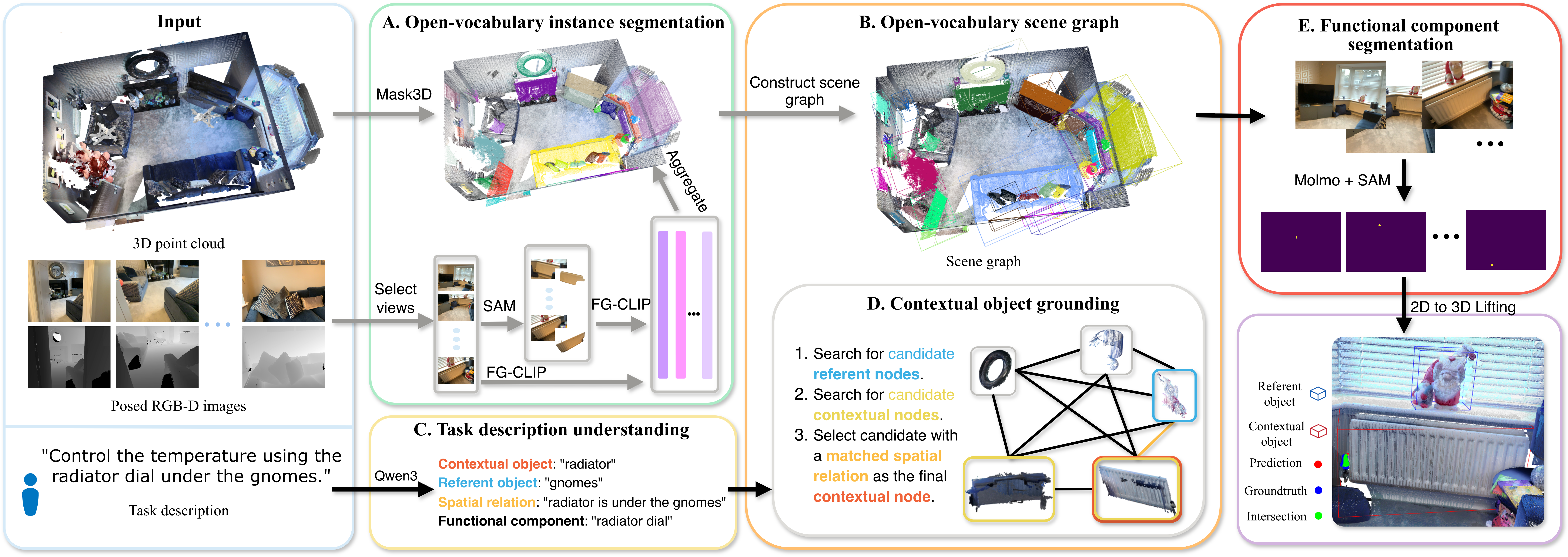}
\caption{T-FunS3D overview: The input of T-FunS3D are posed RBG-images and 3D point clouds of an indoor scene. (A) performs open-vocabulary instance segmentation by associating FG-CLIP~\cite{xie2025fg} visual embeddings to class-agnostic instance segmentation from Mask3D~\cite{schult_mask3d_2023}. We construct a scene graph of the featurized instances (B). Once a task is assigned, we decompose the description into ontologies using Qwen3~\cite{yang2025qwen3} (C) and identify the contextual object in the scene graph (D) by computing the text-visual embedding similarity. Lastly, (E) aggregates 2D masks extracted by combining Molmo~\cite{deitke2025molmo} with SAM~\cite{kirillov2023segment} to obtain 3D segmentation of functional parts.}
\vspace{-.5cm}
\label{fig:pipeline}
\centering
\end{figure*}

\subsection{3D Scene Graph}
3D scene graphs encode objects and spatial concepts (e.g., rooms and floors) as nodes and their relations as edges, offering compact, object-centric representations well suited for robotics~\cite{armeni20193d}.
This decomposition of environments facilitates higher-level reasoning and planning in object-centric tasks. 
Early works like \cite{rosinol2020kimera} and \cite{rosinol20203d} integrate SLAM with annotated data but remain limited to closed sets of objects. 
Harnessing multi-modal large language models, methods like HOV-SG~\cite{werby2024hierarchical} and CLIO~\cite{maggio2024clio} construct hierarchical open-vocabulary 3D scene graphs focusing on high levels of semantic meanings from room over region to object.
Nevertheless, their graph encodes only parent-child relations(e.g., floor-rooms,  room-objects), which is sufficient for navigation but inadequate for interactive tasks.
Such interactive applications in practice require a clear reasoning of relations between objects so the manipulator can infer the target object and its reachable grasps in human-centric spaces.
ConceptGraphs~\cite{gu2024conceptgraphs}, OVSG~\cite{changcontext}, and Open3DSG~\cite{koch_open3dsg_2024} advance this line by encoding open-vocabulary inter-object edges, enabling advance queries including referential expressions. 
However, they still struggle to identify part-level entities required for task execution. 

Contemporary to our work, OpenFunGraph~\cite{zhang2025open} and FunGraph~\cite{rotondi_fungraph_2025} also address the functionality understanding problems using 3D scene graphs.
They both establish nodes and edges for interactive object elements.
Among them, FunGraph includes both inter-object and intra-object edges with respect to the spatial relationships, while OpenFunGraph only models the part-object relationships, specifying the functionality.
Similar to ConceptGraphs~\cite{gu2024conceptgraphs}, these two methods build edges based on multi-view aware natural language descriptions of nodes generated by VLMs and LLMs.
In contrast, instead of using LLM-generated text descriptions, we leverage visual embeddings of multi-view observations of nodes to model the edges between objects, reducing computational and time consumption.
\section{Problem Formulation}

Given an input sequence of posed RGB-D frames $\{\mathcal{I}, \mathcal{D}\}$ and the point cloud $P$ of an indoor environment, the goal is to get a 3D mask that locates the interactive functional component $\mathcal{F}$ in the scene given a manipulation task query~$\mathcal{Q}$.
We assume that the entities captured in the point cloud are static to the extend that the spatial relationships between instances preserve, while local movements are possible.
For example, the coffee table can be moved locally but it is assumed to remain to the right of the sofa and to the left of the radiator.

In this work, we focus on queries that explicitly encompass the information of objects that the robots should interact with, referred as the contextual objects $\mathcal{C}$.
For instance, in ``\textit{Adjust the room's temperature using the radiator thermostat}'', ``radiator'' is the contextual object. 
Frequently, referential descriptions of object locations are given in the queries to reduce ambiguity, such as "\textit{Open the nightstand drawer on the right side of the bed}".
In such type of expressions, the objects used as a reference coordinate are called the referent objects $\mathcal{R}$ (``bed'' in this example).
The corresponding spatial relationships between $\mathcal{C}$ and $\mathcal{R}$ are denoted as 2-tuples $\mathcal{S} = (\mathcal{C}, \mathcal{R})$.
Thus, a query can be generally assumed to be a collection of {\color{Dandelion}\textbf{spatial relations} $\mathcal{S}$}, {\color{CornflowerBlue}\textbf{referent object} $\mathcal{R}$}, {\color{RedOrange}\textbf{contextual object} $\mathcal{C}$}, and its \textbf{functional component} $\mathcal{F}$, i.e.,
$$\mathcal{Q} = \{\mathcal{C}, \mathcal{F}, (\mathcal{S}_1, \mathcal{R}_1, \mathcal{S}_2, \mathcal{R}_2, \dots)\}.$$
Notably, the referent objects and spatial relationships are optional and there can be multiple ones.
The same color codes are adopted in block C and D in Fig.~\ref{fig:pipeline}. 

To tackle this task, we introduce an effective and efficient two-step approach. 
First, we locate the contextual object $\mathcal{C}$ in an open-vocabulary scene graph.
After that, we segment the functional element with respect to the query using the images corresponding to the identified instance of interest.
\section{Open-Vocabulary Functionality Segmentation}
\label{sec:method}

Our proposed pipeline, T-FunS3D, is illustrated in Fig.~\ref{fig:pipeline}.
Given RGB-D observations, ground truth camera poses, and the 3D point cloud of an indoor environment, T-FunS3D provides task-driven hierarchical open-vocabulary functionality segmentation in 3D scenes.
The pipeline consists of four modules organized into two stages.
At the first stage, we perform 3D instance-level open-vocabulary segmentation by extracting semantic embeddings for class-agnostic 3D instance segmentation (Sec.~\ref{subsec:instance segmentation}). 
Then, an open-vocabulary scene graph is constructed based on the featurized 3D instance segments (Sec.~\ref{subsec:scene graph}). 
Note that this stage only needs to be performed once per scene, provided the assumption that the environmental layout remains static.
The second stage begins once a task in free-form text is assigned to the system. 
At this stage, we analyze the task description with a large language model (LLM) to extract the task ontology (Sec.~\ref{subsec:task understanding}). 
Afterwards, the relevant contextual instance is localized in the scene graph (Sec.~\ref{subsec:object grounding}) and we segment its functional components required for task execution (Sec.~\ref{subsec: functionality segmentation}).

\subsection{Open-vocabulary 3D instance segmentation}
\label{subsec:instance segmentation}
To obtain object-level instance proposals in an open-vocabulary setting, we adapt the approach introduced in OpenMask3D~\cite{takmaz_openmask3d_2023}. 
To achieve this, Mask3D~\cite{schult_mask3d_2023} is applied to generate class-agnostic object-level mask proposals and decompose the scene.
To bridge the semantic meaning to the proposals, the semantic features of objects are extracted.
For this purpose, optimal views of the class-agnostic masks are identified by their visibility on various views, which is computed by projecting the associated 3D points on the input RGB.
Once the views with the highest visibility ratio are obtained, semantic features are extracted using the visual encoder of multi-modal language models, inspired by \cite{takmaz_openmask3d_2023}, \cite{nguyen_open3dis_2023}, and \cite{werby2024hierarchical}.
Unlike OpenMask3D, which computes visual embedding using only image crops of the objects, we leverage more pixel information to associate richer semantics with each proposal.
For every extracted instance $P_n$, where $n \in \{1,\dots, N\}$, the top $k \in \{1,\dots,K\}$ views are cropped around the instance at multiple scales.
The $l$-th ($l \in \{1,\dots, L\}$) scale crop of the $k$-th view for the $n$-th instance is denoted as $I_{k, l}^{n}$. 
Subsequently, we compute the open-vocabulary semantics for the instances from both the full-size RGB images and the crops of selected views using FG-CLIP~\cite{xie2025fg}, a multimodal vision-language model.
However, in practice, the target object in views may be occluded or visible only through transparent materials (e.g., windows).
In this case, their crops introduce a foreground-background bias that can reduce the accuracy of visual embeddings dramatically, e.g., the window frame is mistaken for plants outside the window.  
To mitigate this effect, we additionally generate masked crops, denoted as $M_{k, l}^{n}$, where irrelevant pixels are masked out. These masked crops are used as complementary cues for the visual encoder. 
This combination retains contextual information while emphasizing the target instance. 
The final visual embedding for $P_n$ is obtained by averaging across views and scales:
$$f(P_n) = \frac{1}{2K}\sum_k^K \big[f(I_k^n) + \frac{1}{2L}\sum_l^L f(I_{k, l}^{n}) + f(M_{k, l}^{n})\big]$$
where $f(\cdot)$ denotes the visual embedding computation.

\subsection{Open-vocabulary scene graph}
\label{subsec:scene graph}
A scene graph is defined as $\mathcal{G} = (\mathcal{V}, \mathcal{E})$, where $\mathcal{V} = \{v_i\}, i \in \{1, \dots, N\}$ denotes the set of vertices and $\mathcal{E} = \{e_{ij} | e_{ij} = (v_i, v_j), \text{ for } v_i, v_j \in \mathcal{V} \text{ where } i\neq j\}$ the set of edges. 
To enable open-vocabulary querying, we store visual embeddings rather than explicit class labels within the nodes, inspired by \cite{werby2024hierarchical}. 
Edges between two nodes are established by averaging the visual embedding of the full-size images associated with the two vertices. 
In this way, we encode the semantic context surrounding the instances, alongside the inter-instance spatial relations captured across multiple views. 
The edges are computed during querying between all candidate contextual and referent object pairs rather than precomputed and stored in the graph as in previous works like~\cite{gu2024conceptgraphs} (see Sec. \ref{subsec:object grounding}).

\subsection{Task understanding using LLM}
\label{subsec:task understanding}
The task descriptions $\mathcal{Q}$ may involve varying levels of complexity. 
One the one hand, names of the functional entity can be expressed explicitly, like in the query ``Control the temperature using the radiator dial'', or requires to be inferred as for ``Open the door''.
On the other hand, the descriptions may contain no referent objects, and therefore, no spatial relationships, like in the aforementioned two examples.
Thus, basic ontology extraction techniques in natural language processing is insufficient for interpreting free-form task descriptions.

To this end, we leverage a LLM to parse them and extract clues, including spatial relationships $\mathcal{S}$, referent objects $\mathcal{R}$, the contextual object $\mathcal{C}$, and its functional entity $\mathcal{F}$.
However, extracting each category individually can lead to ambiguity.
Take the query ``Open the door next to the window'' and the categories, contextual object and functional entity as an example.
On the one hand, if the LLM is prompted to extract only the contextual object, it may output a generic label such as ``door'', which is ambiguous if multiple doors exist in the environment.
Thus, identifying the target object category is inadequate for the robot to undertake the task.
On the other hand, if the model only extracts the functional object and outputs a ``handle'', it remains unclear whether the agent should manipulate the handle of a door or a window.
Therefore, we prompt a LLM to jointly provide all categories in a single JSON-format output, following the strategy in \cite{corsetti_functionality_2024}. 
Additionally, different from \cite{corsetti_functionality_2024}, we explicitly prompt the LLM to extract the spatial relationships of different instances from the description if they exist.

\subsection{Contextual object grounding}
\label{subsec:object grounding}
The contextual object is retrieved by querying the scene graph based on the task description.
We denote the visual and textual embedding functions by $f(\cdot)$ and $g(\cdot)$, respectively. 
Let $t_{\mathcal{R}}$ and $t_{\mathcal{C}}$ indicate the text descriptions for referent and contextual objects.
After encoding $t_{\mathcal{C}}$ and $t_{\mathcal{R}}$ into textual embeddings using FG-CLIP~\cite{xie2025fg}, we extract candidate nodes for the respective objects based on their embedding similarities.
For contextual object $\mathcal{C}$, we retrieve candidate nodes $\{C_i\}$ from the vertices $\mathcal{V}$ whose visual embeddings are most similar to the textual embedding of~$\mathcal{C}$:
$$C_i = \arg \max_{v_{i} \in \mathcal{V}} \: sim(f(v_{i}), g(t_{\mathcal{C}}))$$
where $sim(\cdot\,, \cdot)$ computes the cosine similarity of two embedding vectors.
Analogously, if there are referent objects $\mathcal{R}$ specified in $\mathcal{Q}$, we continue to select candidate nodes $\{R_j\}$ associated with $t_{\mathcal{R}}$:
$$R_j = \arg \max_{v_{j} \in \mathcal{V}} \: sim(f(v_{j}), g(t_{\mathcal{R}}))$$

Once we obtain the candidates, we examine the edges between all pairs $\{S_{ij}\} = \{(C_i, R_j)\}$ to identify those that have high embedding similarity scores with the spatial relationships $\mathcal{S}$ described in the query.
Here, we denote the text description of spatial relationships by $t_\mathcal{S}$.
For every extracted spatial relations involving the contextual object, the computation is formulated as:
$$S = \arg \max_{s \in S_{ij}} \: sim(f(s), g(t_\mathcal{S}))$$
The nodes in the corresponding candidate pairs with valid $S$ are anchored as the contextual object and referent objects. 
If there are no spatial relationships in the descriptions, the best candidate contextual node is fed to the next module.

\subsection{Functional component segmentation}
\label{subsec: functionality segmentation}
We segment the functional components of the contextual object by combining the VLM Molmo~\cite{deitke2025molmo} with SAM~\cite{kirillov2023segment}, inspired by Fun3DU~\cite{corsetti_functionality_2024}. 
Once the contextual object is identified, we retrieve its corresponding selected views and feed them into Molmo. 
Specifically, the task description ontology extracted in Sec. \ref{subsec:task understanding} is leveraged as a prompt to guide Molmo in the functional element detection. 
As a result, Molmo outputs pixel coordinates corresponding to the functional component of interest. 
These pixel coordinates further serve as prompt for SAM to generate the 2D segmentation masks. 
By default, multiple masks are extracted for every prompt point in SAM.
Each mask is delivered along with a confidence score, i.e., the estimated intersection over union.
However, in practice, we notice that the masks with the highest scores tend to segment the entire object instead of the part indicated by the prompts, especially for crops with lower resolution.
Therefore, instead of the masks with the highest confidence score, we resort to the smallest masks among the predictions.
This design choice is further justified by an ablation study in Sec.~\ref{Sec:ablation study}.
The predicted 2D masks are subsequently back-projected into the 3D space, where multi-view aggregation of these lifted masks yields the final 3D functionality segmentation.
\section{Experiments}
\subsection{Experiment setup}
\subsubsection{Implementation}
T-FunS3D is a training-free method leveraging several pre-trained models. 
We build a scene-scale open-vocabulary instance segmentation (A in Sec.~\ref{fig:pipeline}) based on OpenMask3D~\cite{takmaz_openmask3d_2023} with no retraining but several modifications on its pipeline.
Their class-agnostic instance proposal model ~\cite{schult_mask3d_2023} is trained on ScanNet200~\cite{rozenberszki2022language}, whose point clouds are roof-less and have fewer points per scene compared to SceneFun3D~\cite{delitzas2024scenefun3d}.
Therefore, we preprocess the point clouds  
to fit them in the range and form on which the proposal model is trained (see supplementary materials).
In addition, we only preserve class-agnostic masks of the highest confidence scores, while OpenMask3D maintains all masks generated by the model.
This is based on the observations that more than half of the proposed masks identify no meaningful objects or duplicate each other, resulting in unnecessary consumption in the visual embedding computation.
Another deviation from OpenMask3D is that we compute visual features of the proposed instance from richer sources, including the full-size RGB, crops around the instance, and masked crops of the top-$k$ views associated to this instance, described in Sec.~\ref{subsec:instance segmentation}.
This combination provides not only more reliable visual information about instances themselves, preventing foreground-background-bias, but also more context information of the surroundings, facilitating referring localization of objects.
This design choice is justified in the Sec.~\ref{Sec:ablation study}.
In our standard method, we sample one from every five frames of the complete video as input and set $k$ to 5.

Multi-modal language models, such as CLIP~\cite{radford2021learning}, SigLIP~\cite{zhai2023sigmoid}, and SigLIP2~\cite{tschannen2025siglip}, that offer both visual and textual encoders, are the core component to bridge semantic meaning to 3D scenes.
For visual and textual embedding computation, FG-CLIP~\cite{xie2025fg} is applied because of its reliable performance in preserving the fine-grained details.

When a query is given, QWen3-14B~\cite{yang2025qwen3} is employed to analyze the task descriptions. Specifically, we disable the thinking capability of the model and query in multi-turn conversations to allow faster inference without sacrificing the output accuracy. Eventually, the functional part segmentation is achieved by combining Molmo-7B-D~\cite{deitke2025molmo} and SAM~\cite{kirillov2023segment}.

\subsubsection{Dataset}
We evaluated our method on the SceneFun3D~\cite{delitzas2024scenefun3d} dataset. 
It is the only available dataset that supports task-driven functionality segmentation in 3D by providing annotations for functional components together with a diverse set of tasks for indoor scenes.
In this paper, we reported the evaluation results on the validation split of this dataset, which contains $30$ scenes with $445$ task descriptions.

\begin{figure}[ht]
\centering
\includegraphics[width=\linewidth]{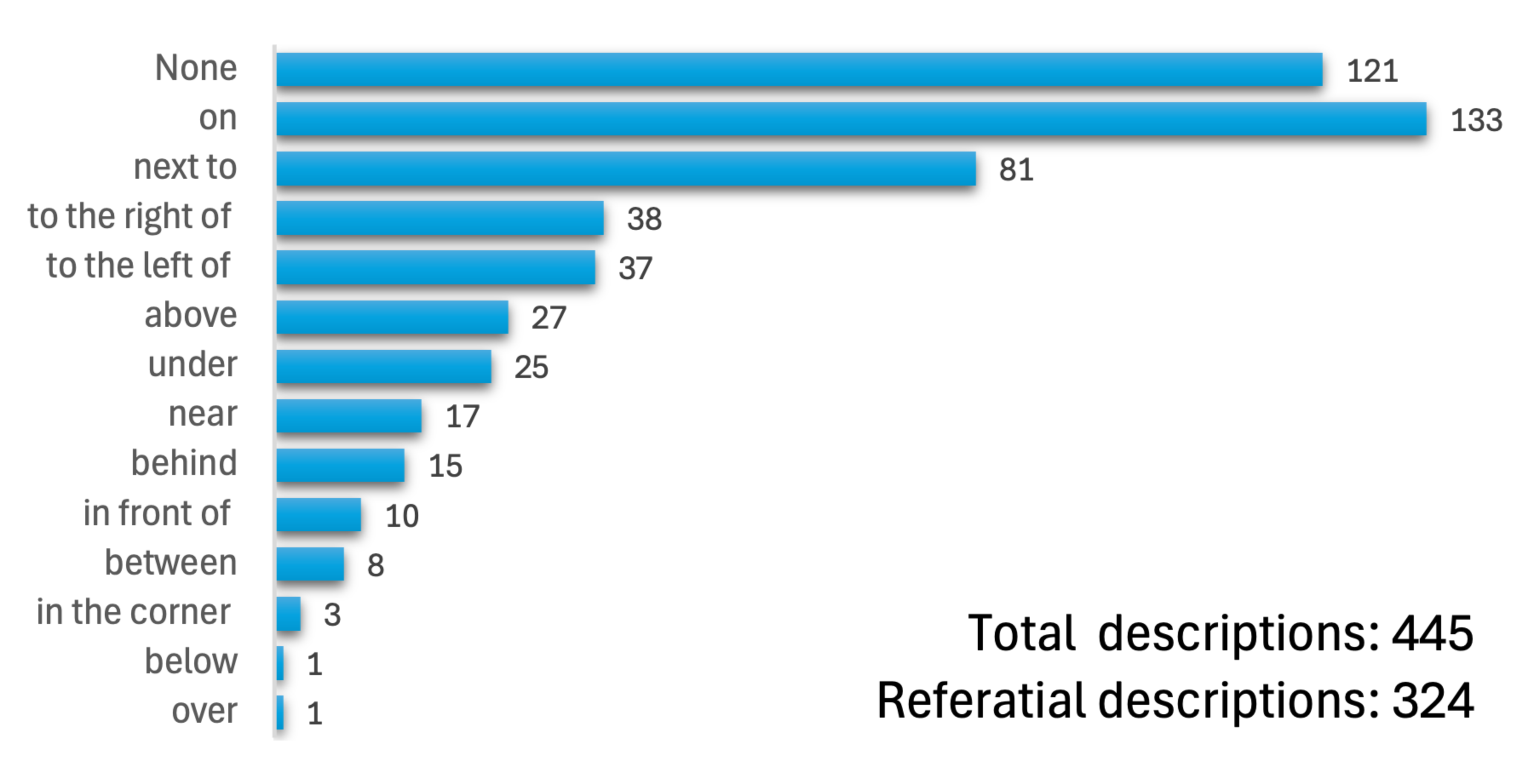}
\caption{Distribution of spatial relationships in the task descriptions provided in the validation split of the SceneFun3D dataset~\cite{delitzas2024scenefun3d}.}
\label{fig:relation distribution}
\vspace{-.3cm}
\centering
\end{figure}

Among the task descriptions in this split, about three quarters possess referring expressions, as shown in Fig.~\ref{fig:relation distribution}.
Note that the total number of counted spatial relationships ($121$ of ``None'' and $396$ of the others) does not match the number of descriptions, as some descriptions attribute multiple relationships. 
More experiment results can be found on the supplementary materials.

\subsubsection{Evaluation metrics}
As suggested in SceneFun3D~\cite{delitzas2024scenefun3d}, we measure different metrics, including the Average Precision (AP) at Intersection over Union (IoU) thresholds of $0.25$ (AP$_{25}$) and $0.5$ (AP$_{50}$), and the mean Average Precision (mAP) averaging over IoU thresholds from $0.5$ to $0.95$ with a step of $0.05$. In addition, like ~\cite{corsetti_functionality_2024}, we also report the mean IoU and the Average Recall (AR), including AR$_{25}$, AR$_{50}$, and mAR of the same conditions as AP.

\subsubsection{Baselines}
For comparison, we use Fun3DU~\cite{corsetti_functionality_2024} as the main baseline method on functionality segmentation tasks. 
In addition, due to the lack of open-sourced methods dedicated to fine-grained functional component segmentation at the time of writing, T-FunS3D is also compared against state-of-the-art open-vocabulary 3D instance segmentation, including OpenMask3D~\cite{takmaz_openmask3d_2023},  OpenIns3D~\cite{huang_openins3d_2023}, and LeRF~\cite{kerr2023lerf}.
To ensure consistency, we report the results of the baseline methods presented in~\cite{corsetti_functionality_2024}.
Additionally, for a fair comparison, we reproduced the results of a primary baseline, Fun3DU~\cite{corsetti_functionality_2024}, using their released code and the same task parsing inputs as T-FunS3D, denoted as Fun3DU\textdagger{} in the tables~below.

\subsection{Results}
\subsubsection{Functionality segmentation}

TABLE~\ref{tab:result1} presents the evaluation on the performance of functional part segmentation.
As shown, OpenMask3D scores nearly zero precision and low recall due to the increased complexity of the tested scenes compared to its training data.
The other two instance segmentation methods score high recall but zero precision, indicating that they tend to segment the entire object rather than the fine-grained object parts.
Fun3DU~\cite{corsetti_functionality_2024} achieves good figures for all metrics, demonstrating reliable performance.
However, it can not be neglected that Fun3DU performs contextual object detection on all 2D views using an additional VLM, which requires repetition for new queries.
Despite that, T-FunS3D still surpasses Fun3DU and also the other baselines with the highest precision and IoU, as well as generally high recall.
In comparison, T-FunS3D demonstrates strong functionality segmentation performance with at least {\color{Green}$+1.2$} AP$_{25}$ and {\color{Green}$+0.5$} mIoU improvement over baseline methods. The improvement increases to {\color{Green}$+11.1$} AP$_{25}$ and {\color{Green}$+3.7$} mIoU compared to the reproduced results of Fun3DU.
\begin{table}[htbp]
\vspace{-.4cm}
\setlength\tabcolsep{2pt}
\centering
\caption{Performance comparison on SceneFun3D~\cite{delitzas2024scenefun3d} dataset.}
\label{tab:result1}
{\renewcommand{\arraystretch}{1.2}
\begin{tabular*}{.95\columnwidth}{@{\extracolsep{\fill}} l|ccc|ccc|c}
\hline
Methods     & mAP & AP50 & AP25 & mAR  & AR50 & AR25 & mIoU \\ \hline
OpenMask3D~\cite{takmaz_openmask3d_2023} & 0.2 & 0.2  & 0.4  & 20.3 & 24.5 & 27   & 0.2  \\
OpenIns3D~\cite{huang_openins3d_2023}  & 0   & 0    & 0    & \textbf{40.5} & \textbf{46.7} & \textbf{51.5} & 0.1  \\
LERF~\cite{kerr2023lerf}       & 0   & 0    & 0    & 34.2 & 35.1 & 36   & 0    \\
Fun3DU~\cite{corsetti_functionality_2024}     & 7.6 & 16.9 & 33.3 & 27.4 & 38.2 & 46.7 & 15.2 \\ \hline
Fun3DU~\cite{corsetti_functionality_2024}\textdagger     & 4.4 & 10.3 & 23.4 & 30.9 & 42.3 & 49.7 & 12.0 \\ \hline
T-FunS3D (ours)     & \textbf{8.1} & \textbf{17.8} & \textbf{34.5} & 23.8 & 35.8 & 46.9 & \textbf{15.7} \\
\hline
\end{tabular*}}
\end{table}

\begin{figure*}[ht]
\centering
\vspace{.1cm}
\includegraphics[width=\linewidth]{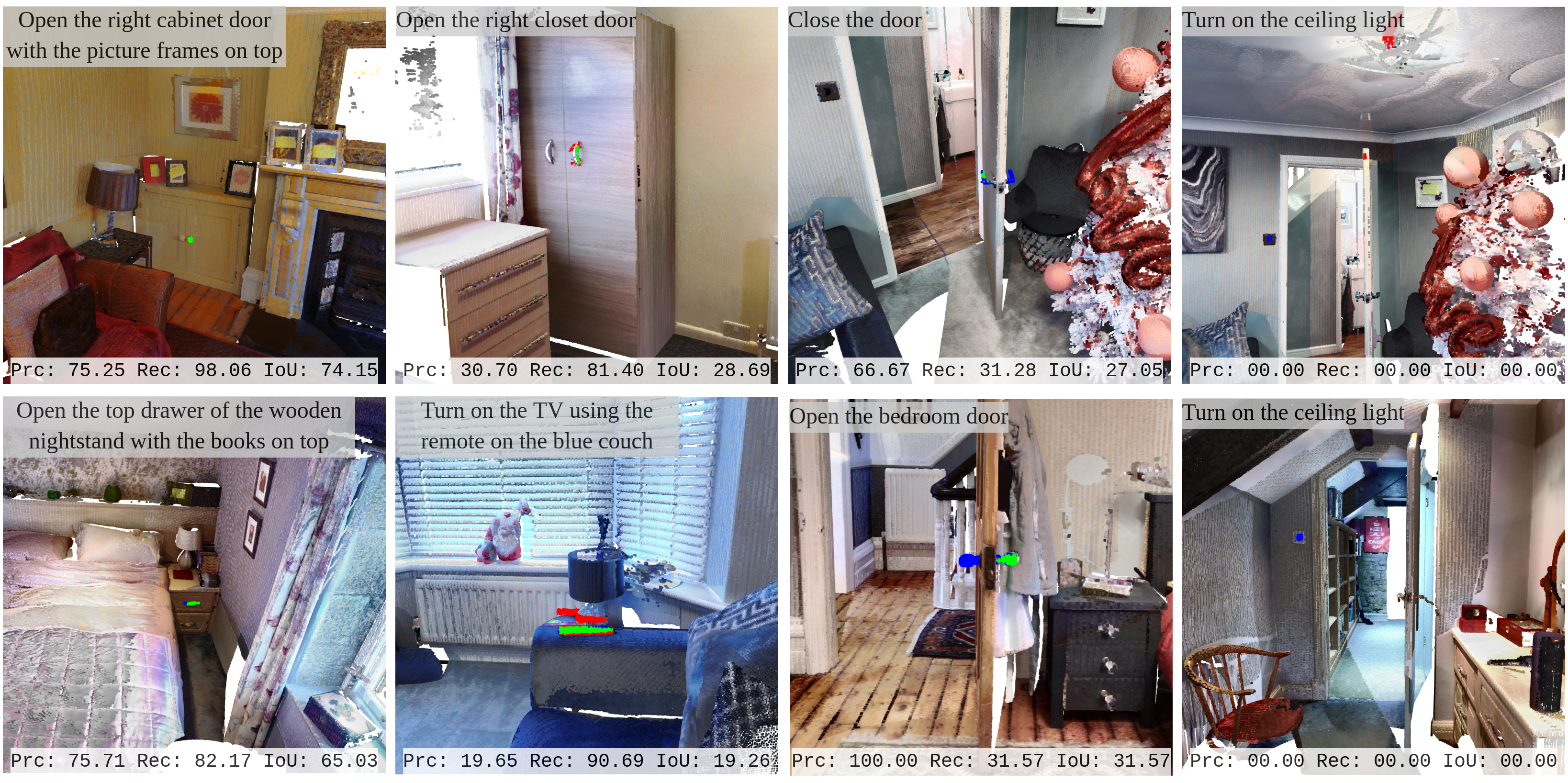}
\caption{Qualitative examples of T-FunS3D. We visualize point clouds around the functional components: red points indicate predictions, blue points denote ground truth, and green points represent overlaps.}
\label{fig:qualitative_results}
\vspace{-.5cm}
\centering
\end{figure*}

\subsubsection{Referring grounding}
\begin{table}[htbp]
\setlength\tabcolsep{2pt}
\centering
\vspace{.15cm}
\caption{Performance on task descriptions with spatial referring expressions in the validation split of SceneFun3D~\cite{delitzas2024scenefun3d}.}
\label{tab:referring results}
{\renewcommand{\arraystretch}{1.2}
\begin{tabular*}{.9\columnwidth}{@{\extracolsep{\fill}} l|ccc|ccc|c}
\hline
Methods              & mAP & AP50 & AP25 & mAR  & AR50 & AR25 & mIoU \\ \hline
Fun3DU~\cite{corsetti_functionality_2024}\textdagger  &   3.83 & 9.26 & 22.53 & \textbf{30.77} & \textbf{41.36} & \textbf{48.77} & 11.73 \\
T-FunS3D (ours)  &    \textbf{8.11} & \textbf{19.20} & \textbf{34.98} & 25.17 & 37.46 & 47.06 & \textbf{16.24}   \\ \hline 
\end{tabular*}}
\vspace{-.3cm}
\end{table}
TABLE~\ref{tab:referring results} compares T-FunS3D and the baseline method, Fun3DU~\cite{corsetti_functionality_2024}, on a subset of the validation split in SceneFun3D~\cite{delitzas2024scenefun3d} which consists of descriptions that identify the contextual object's location by referring to its surrounding items (e.g., ``Turn on the lamp on the side table next to the Christmas tree''), as interpreted in Fig.~\ref{fig:relation distribution}.
Results on this subset reflect the capability of localizing objects in 3D scenes from such complex referring expressions. 
As shown in the TABLE~\ref{tab:referring results}, T-FunS3D achieves much higher precision and IoU scores in localizing the functional interactive elements than Fun3DU, showing at least {\color{Green}$+12.45$} AP$_{25}$ and {\color{Green}$+4.51$} mIoU improvement. 
This better performance is due to two core design choices in T-FunS3D. 
First, T-FunS3D decomposes the input query and extracts the spatial relations using the LLM, which serves in the subsequent stage as prompts to query on the concrete instances.
The extraction of relations makes the prompt more precise and context-aware, increasing the likelihood of accurately locating the target object.
Second, the scene graph in T-FunS3D encodes the inter-object spatial relations, which ease the object localization based on referential descriptions.
T-FunS3D scores slightly lower recall compared to Fun3DU due to the use of the smallest 2D masks from SAM rather than the ones with the highest confidence, as in Fun3DU.
A detailed discussion on the design choice is in Sec.~\ref{Sec:ablation study}

\subsubsection{Qualitative results}
Fig.~\ref{fig:qualitative_results} presents qualitative examples of our functionality segmentation, intuitively illustrating both the capabilities and the current limitations of T-FunS3D.
We aggregate 2D segmentation masks obtained from selected views based on contextual object's visibility onto the 3D point cloud. 
However, images captured from large oblique viewing angles can lead to less accurate segmentation, as indicated by the low precision and recall metrics in the second and third columns in Fig.~\ref{fig:qualitative_results}.
Additionally, our method exhibits limitations when functional elements lack physical attachment to the contextual objects.
For instance, it fails to segment the ceiling light switch (yielding zero IoU) in the rightmost column.

\subsection{Runtime comparison}
TABLE~\ref{tab:runtime} compares the runtime performance of T-FunS3D against two baselines for instance \raisebox{.5pt}{\textcircled{\raisebox{-.9pt} {1}}} and functionality segmentation \raisebox{.5pt}{\textcircled{\raisebox{-.9pt} {2}}}. 
The reported runtimes are experimentally measured on an NVIDIA A40.
Considering the full pipeline, T-FunS3D achieves slightly higher accuracy than the state-of-the-art Fun3DU~\cite{corsetti_functionality_2024} in a much shorter time.
Unlike Fun3DU, which segments instances of interest across all images, T-FunS3D and OpenMask3D generate class-agnostic instance proposals from point clouds (\raisebox{.5pt}{\textcircled{\raisebox{-.9pt} {a}}}) and compute embeddings only from views where each instance is most visible (\raisebox{.5pt}{\textcircled{\raisebox{-.9pt} {b}}}). 
T-FunS3D is faster than OpenMask3D at \raisebox{.5pt}{\textcircled{\raisebox{-.9pt} {1}}} because T-FunS3D retains only high-confidence proposals. 
A comparison with OpenMask3D at \raisebox{.5pt}{\textcircled{\raisebox{-.9pt} {2}}} is not applicable, as OpenMask3D does not perform functionality segmentation.
T-FunS3D further reduces runtime at stage \raisebox{.5pt}{\textcircled{\raisebox{-.9pt} {2}}} by reusing the cached views, which are less than the required images in Fun3DU.
As a reference, Fun3DU processes one task at \raisebox{.5pt}{\textcircled{\raisebox{-.9pt} {2}}} averagely in 118.4 seconds on an NVIDIA A100 as reported in \cite{corsetti_functionality_2024}.
Moreover, for every new query, T-FunS3D requires shorter process time by caching object visual embeddings with top-ranked views obtained at \raisebox{.5pt}{\textcircled{\raisebox{-.9pt} {1}}}, whereas Fun3DU must repeat the complete segmentation pipeline.
\begin{table}[htbp]
\centering
\caption{Runtime comparison at instance and functionality segmentation}
\label{tab:runtime}
{\renewcommand{\arraystretch}{1.2}
\begin{tabular}{ l| c | c }
\hline
Methods     & \raisebox{.5pt}{\textcircled{\raisebox{-.9pt} {1}}} OV Inst. Segm. & \raisebox{.5pt}{\textcircled{\raisebox{-.9pt} {2}}} Func. Segm.  \\
 &  (per-scene average) &  (per-query average) \\
\hline
OpenMask3D~\cite{takmaz_openmask3d_2023}& \raisebox{.5pt}{\textcircled{\raisebox{-.9pt} {a}}} 30s + \raisebox{.5pt}{\textcircled{\raisebox{-.9pt} {b}}} 720s               & -                          \\
Fun3DU~\cite{corsetti_functionality_2024}    & 1920s               & 167s               \\
    T-FunS3D (ours)& \raisebox{.5pt}{\textcircled{\raisebox{-.9pt} {a}}} \textbf{12s} + \raisebox{.5pt}{\textcircled{\raisebox{-.9pt} {b}}} \textbf{580s}        & \textbf{78s}  \\
\hline
\end{tabular}}
\vspace{-.3cm}
\end{table}

\subsection{Ablation study}
\label{Sec:ablation study}
We assess the impact of two methodological design choices introduced in Sec.~\ref {sec:method}, i.e., visual embedding computation with richer information and aggregation of the smallest SAM masks into 3D, on the final performance.
Experiments in this ablation study are conducted on the first $10$ visits in the validation split of the SceneFun3D~\cite{delitzas2024scenefun3d} dataset.
The results in TABLE~\ref{tab:ablation study} emphasize the importance of these designs.

As the primary comparison baseline, we report the reproduced results of Fun3DU~\cite{corsetti_functionality_2024} under the same task understanding condition as ours in row 1 (\textdagger).
In rows 2 and 3, we replace the collection of images, including RGB images in full size, crops, and masked crops, with varying combinations.
\begin{table}[htbp]
\setlength\tabcolsep{2pt}
\centering
\caption{Ablation study on T-FunS3D design choices for functionality segmentation on a subset of the validation split in SceneFun3D~\cite{delitzas2024scenefun3d}}
\label{tab:ablation study}
{\renewcommand{\arraystretch}{1.2}
\begin{tabular*}{.95\columnwidth}{@{\extracolsep{\fill}} l|ccc|ccc|c}
\hline
Methods              & mAP & AP50 & AP25 & mAR  & AR50 & AR25 & mIoU \\ \hline
Fun3DU~\cite{corsetti_functionality_2024}\textdagger &  4.41 & 10.49 & 22.38 & 30.28 & 41.26 & 49.65 & 11.07 \\ \hline
Ours w/o FI    & 5.73 & 13.99 & 26.57 & 22.80 & 33.57 & 41.26 & 13.40 \\ 
Ours w/o MC    & 5.53 & 17.02 & 29.08 & 21.84 & 28.37 & 36.17 & 13.37 \\
Ours w/o SM    & 4.46 & 10.36 & 23.42 & 30.95 & 42.34 & 49.77 & 12.03 \\ \hline
Ours standard  & 6.41 & 15.49 & 31.69 & 23.03 & 32.39 & 40.14 & 14.98 \\ \hline 
\end{tabular*}}
\vspace{-.3cm}
\end{table}
Row 2 (w/o FI) ignores full-size images in visual embedding computation, while row 3 (w/o MC) combines full-size images and crops, but discards the masked crops.
In row 4 (w/o SM), we explore the use of 2D masks produced by SAM~\cite{kirillov2023segment} with the highest score for every prompt point instead of the smallest ones.
All configurations exhibit general performance declines compared to our standard method, except the method in row 1 achieves higher recalls.
The abundant pixel information fed into the embedding computation contributes around {\color{Green}$+0.8$} mAP and {\color{Green}$+1.5$} mIoU, while leveraging the smallest masks generated by SAM leads to about {\color{Green}$+2.0$} mAP and nearly {\color{Green}$+3.0$} mIoU improvement. 
The drop of recall from row 4 to row 5, implying more false negatives in the prediction, indicates, in turn, a higher concentration on the functional entities by lifting the smallest 2D masks to 3D.
\section{Conclusion}
We presented T-FunS3D, a novel training-free method for task-driven functionality segmentation in 3D scenes leveraging open-vocabulary scene graphs and VLMs.  
As its core, the proposed open-vocabulary scene graph represents nodes and edges with semantic embeddings, enabling T-FunS3D to efficiently localize task-relevant objects and achieve accurate functionality segmentation in a hierarchical manner.
In future work, fault-handling mechanisms could be integrated between modules to enhance the overall robustness of the pipeline. 
We hope this work inspires continued advancement in open-vocabulary functionality segmentation, facilitating the flexible and efficient grounding of task-relevant scene entities for interactive robotic applications in real-time.

\bibliographystyle{IEEEtran}
\bibliography{IEEEtranBST/ref}

\clearpage
\makeatletter
\def\@seccntformat#1{\csname the#1\endcsname\quad}
\def\@seccntformat#1{%
  \csname the#1\endcsname
  \hspace{0.3em}}   
\renewcommand\paragraph{\@startsection{paragraph}{4}{0.8em}%
                                    {0.5ex plus .1ex minus .1ex}%
                                    {0.0em}%
                                    {\normalfont\normalsize\bfseries}}
\makeatother

{
\newpage
   \twocolumn[
    \centering
    \Large
    \textbf{T-FunS3D: Task-Driven Hierarchical Open-Vocabulary 3D Functionality Segmentation}\\
    \vspace{0.5em}Supplementary Material \\
    \vspace{1.0em}
   ] 
}

\setcounter{figure}{0}
\setcounter{table}{0}
\setcounter{section}{0}
\setcounter{subsection}{0}
\setcounter{paragraph}{0}
\renewcommand{\thefigure}{A\arabic{figure}}
\renewcommand{\thetable}{A\arabic{table}}
\renewcommand{\thesection}{\Roman{section}}
\renewcommand{\thesubsection}{\thesection.\arabic{subsection}}
\renewcommand{\theparagraph}{}

\section{Additional experimental results}


In addition to the results on the validation split reported in our paper, we also conducted evaluation on training split of the SceneFun3D~\cite{delitzas2024scenefun3d} dataset. 
The training split contains $200$ scenes with $2598$ task descriptions. 
The scenes in this split are in general more complex and have a larger number of point clouds, averaging about $10$ million points, while the point clouds in the validation split are limited to about $5$ million points on average.
Therefore, the training split poses additional challenges on our method that proposes instance segmentation based on point cloud input. 
We alleviate the complexity by performing a preprocess that combines downsampling and strategic filtering, detailed in Sec.~\ref{Sec:implementation details}.

\begin{figure}[htbp]
\centering
\includegraphics[width=0.95\columnwidth]{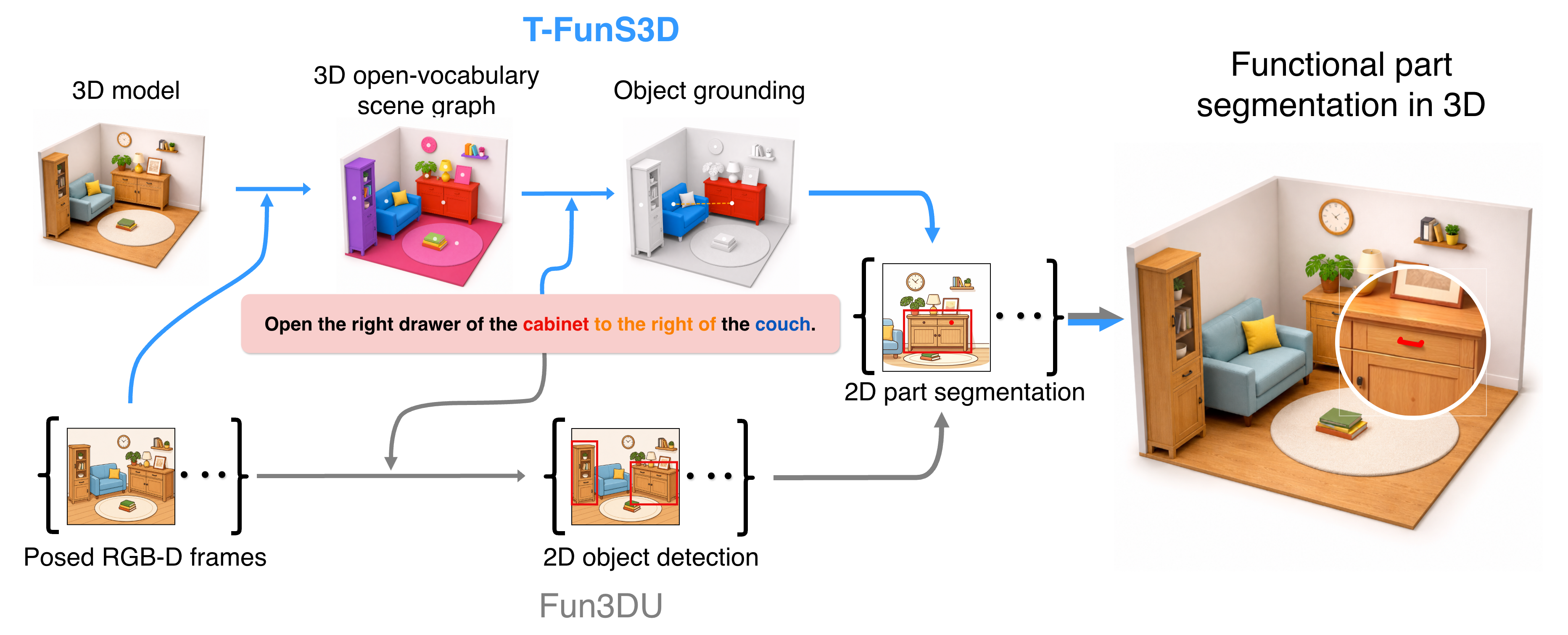}
\caption{Comparing the pipelines of T-FunS3D and its baseline Fun3DU\cite{corsetti_functionality_2024}}%
\label{fig:pipeline comparison}
\end{figure}

The evaluation on the training split is showed in Table~\ref{tab:training_split}.
Analogous to Tab.~\ref{tab:result1} in the paper, we reference the baseline results including OpenMask3D~\cite{takmaz_openmask3d_2023}, OpenIns3D~\cite{huang_openins3d_2023}, LERF~\cite{kerr2023lerf}, and Fun3DU~\cite{corsetti_functionality_2024} reported in \cite{corsetti_functionality_2024}. Additionally, we reproduced Fun3DU~\cite{corsetti_functionality_2024} using the same task description parsing output and the results are report in the row marked with \textdagger.
All instance segmentation methods including OpenMask3D~\cite{takmaz_openmask3d_2023}, OpenIns3D~\cite{huang_openins3d_2023}, and LERF~\cite{kerr2023lerf} exhibit the same pattern as the evalution in Tab.~\ref{tab:result1}; e.g., they tend to segment a whole object though only a interactive part is asked.
OpenMask3D~\cite{takmaz_openmask3d_2023} relies on a 3D encoder~\cite{schult_mask3d_2023}, thus it is mostly affected by the increased complexity in this data split.
In contrast, although our method is implemented based on the same 3D encoder, T-FunS3D does not show a obvious performance decline, which implies the effectiveness of the preprocess approach.
Fun3DU~\cite{corsetti_functionality_2024} is robust to the higher scene complexity, since it is an image-based method, performing segmentation on 2D views.
Nevertheless, Tab.~\ref{tab:training_split} demonstrates that T-FunS3D is still able to achieve performance comparable to Fun3DU~\cite{corsetti_functionality_2024}.
Moreover, in comparison, T-FunS3D achieves faster inference due to hierarchical segmentation strategy based on open-vocabulary scene graphs. A pipeline comparison between T-FunS3D and Fun3DU~\cite{corsetti_functionality_2024} is illustrated in Fig.~\ref{fig:pipeline comparison}.

Interactive visualization is available on our \href{https://t-funs3d.github.io}{project page}.

\begin{table}[htbp]
\setlength\tabcolsep{2pt}
\centering
\caption{Performance comparison on training split in the SceneFun3D~\cite{delitzas2024scenefun3d} dataset.}
\label{tab:training_split}
{\renewcommand{\arraystretch}{1.2}
\begin{tabular*}{.95\columnwidth}{@{\extracolsep{\fill}} l|ccc|ccc|c}
\hline
Methods              & mAP & AP50 & AP25 & mAR  & AR50 & AR25 & mIoU \\ \hline\hline
OpenMask3D~\cite{takmaz_openmask3d_2023} & 0.0 & 0.0  & 0.0  & 1.2  & 1.4  & 2.6  & 0.1  \\
OpenIns3D~\cite{huang_openins3d_2023}  & 0.0 & 0.0  & 0.0  & \textbf{32.3} & \textbf{37.1} & 39.9 & 0.0    \\
LERF~\cite{kerr2023lerf}      & 0.0 & 0.0  & 0.0  & 23.9 & 24.6 & 25.1 & 0.0    \\
Fun3DU~\cite{corsetti_functionality_2024}              & 6.1 & 12.6 & 23.1 & 23.9 & 32.9 & 40.5 & 11.5 \\ \hline \hline
Fun3DU~\cite{corsetti_functionality_2024}\textdagger     & 5.3	& 11.0	& 21.8	& 28.0	& \textbf{37.1}	& \textbf{45.0}	& 11.2 \\ 
T-FunS3D (ours)     & \textbf{7.1}	& \textbf{13.6}	& \textbf{24.9}	& 19.1	& 28.8	& 38.4	& \textbf{11.7}   \\ \hline
\end{tabular*}}
\end{table}

\section{Implementation Details}
\label{Sec:implementation details}
\paragraph{Preprocessing}
We perform preprocessing steps on the input point cloud to align them with the point cloud format that our instance segmentation proposal module~\cite{schult_mask3d_2023}~\cite{takmaz_openmask3d_2023} is trained on. There are two distinctions between the train data and the data in SceneFun3D~\cite{delitzas2024scenefun3d}:
\begin{itemize}
    \item The point clouds in SceneFun3D~\cite{delitzas2024scenefun3d} contains points that either do not belong to the task area or are caused by reflectance.
    \item SceneFun3D~\cite{delitzas2024scenefun3d} captures point cloud with the ceiling, which does not exist in the train data of Mask3D\cite{schult_mask3d_2023}. 
\end{itemize}
These are properties lead to redundant and noisy point clouds, which are detrimental to obtaining a clean and compact segmentation proposals using Mask3D~\cite{schult_mask3d_2023}.

Before addressing these two issues, we first downsample the point cloud, shrinking the number of points to one-third of the original.
After that, DBSCAN filtering is applied to retain only the largest portion of the point cloud based on the fact that reflected points and points that belong to other area such as other rooms or street views are normally spatially not connected to the main task region.
To remove the ceiling points, we trivially filter the points with the highest z coordinates.

\paragraph{Scene graph construction}
In contrast to conventional scene graph construction, our graph contains merely nodes after construction, while edges are established during graph searching. This strategy provides benefits that the complexity of the graph will not be expanded exponentially with the increasing node number. Instead, once the candidate nodes are extracted during querying, their mutual edges can be computed by averaging the feature embeddings of the full-sized images of the corresponding two vertices.
Specifically, let's assumed a list of nodes $R = \{r_{1}, r_{2}, \dots, r_{k}\}$ and another list $C = \{c_{1}, c_{2}, \dots, c_{k}\}$ are extracted as candidate referent and contextual objects, respectively. We denote the a set of $n$ selected full-sized images for one object as $I^{node} = \{I^{node}_{1}, I^{node}_{2}, \dots, I^{node}_{n}\}$. Then, we compute the semantic embeddings between $r_{k}$ and $c_{k}$ as $$e_{r_{k}, c_{k}} = \frac{1}{n}\big(\sum\limits_{i = 1}^{n} f(I^{r_{k}}_{i}) + \sum\limits_{i = 1}^{n} f(I^{c_{k}}_{i})\big)$$
The node with the edge that matches the text description of the queried spatial relationships is then considered as the correct response.

We do not handle the ambiguity explicitly in this work, instead we consider the top-$k$ obtained nodes as reasonable candidates and feed their images to a VLM for part segmentation. The final 3D segmentation is obtained based on a voting schema, which potentially exclude incorrect nodes.

The edges in the scene graph are not defined based on the visual similarity of the full sized images. Instead, they are computed as the average visual embeddings of their vertices. If two objects are observed from the same image, this image will contribute twice to the final average visual embedding stored in the edge, implicitly stressing the relations between these two objects. The similarity scores between the edges (visual embedding of spatial relations) and the text description of the spatial relations (textual embedding) is used to select the correct instance to complete the assigned task.
The edges defined in such way provide semantic information to identify the nodes that fulfill the queried relations. However, they cannot act effectively as the pose of images, as the metric information is included in the embeddings.




\section{Scaling to New Environment and to Environmental Changes}
The runtime of instance segmentation, particularly, associating class-agnostic instance segmentation with open-vocabulary embeddings is proportional to the number of objects in a scene because the association is conducted for every segmented instance individually.
On the one hand, in this work, we assume that the robot has captured the scene completely and the camera poses and point cloud reconstruction of the scene have been obtained by running SLAM or SfM algorithms. These data are, therefore, considered input to our pipeline. Incremental observations are not our research scope in this work. On the other hand, if the new observations are taken or the environment changes, one could update the scene graph by modifying nodes and edges, respectively, rather than rebuild the graph from scratch.

\section{Discussion and Limitations}

We introduced a stage-wise pipeline leveraging recent developments in large models and the scene graph representation to separate the coarse environmental understanding and the fine task-relevant object and part localization in two steps. 
Even though T-FunS3D requires shorter runtime and less memory footprints than its main competitor, approaching usability in real-world applications, it is still limited to the reliability on computational capacities required by running LLMs.

In addition, although components could be upgraded to improve the overall performance, the multi-stage design may suffers from cumulative error propagation across its components.  
For example, a part of failure cases stems from the missed segmentation of small objects, such as light switches and sockets, in the instance segmentation phase. 
One future work is to cooperate mechanisms to detect and handle the accumulative errors to improve robustness.

Furthermore, the final segmentation is still heavily relied on the observation views, although using voting schema can alleviate the problem of over-floated masks. 
We expect that this limitation can be resolved in real robotic applications incorporating active exploration since the robots can maneuver around the contextual object to obtain more observations.

Moreover, our method assumes a observable physical connection between objects and their interactive components. 
However, remote relationships that are not easy to judge based on merely visual cues, such as those between ceiling light and its switch, are difficult to determine.
Further investigation is needed to generalize our method for these remote relationships, e.g., the confidence-aware approach in~\cite{zhang2025open}.

\end{document}